\pdfoutput=1
\PassOptionsToPackage{table}{xcolor}
\documentclass{article} 
\usepackage{iclr2027_conference,times}
\iclrfinalcopy

\usepackage{amsmath,amsfonts,bm}

\def\eqref#1{equation~\ref{#1}}

\def\1{\bm{1}}

\DeclareMathAlphabet{\mathsfit}{\encodingdefault}{\sfdefault}{m}{sl}
\SetMathAlphabet{\mathsfit}{bold}{\encodingdefault}{\sfdefault}{bx}{n}

\usepackage{url}
\usepackage{graphicx}
\usepackage{wrapfig}
\usepackage{algorithm}
\usepackage{algorithmic}
\usepackage{booktabs}
\usepackage{multirow}
\usepackage{colortbl}
\usepackage[table]{xcolor}
\usepackage{enumitem}
\SetEnumitemKey{apptoc}{itemsep=0pt plus 1fil, topsep=0pt plus 1fil, parsep=0pt, partopsep=0pt}
\usepackage[hidelinks]{hyperref}
\usepackage[most]{tcolorbox}
\definecolor{aecggreen}{RGB}{226,240,217}
\definecolor{gainred}{HTML}{B80000} 
\definecolor{lossteal}{RGB}{42,157,143}

\newcommand{\gain}[1]{\hspace{0.1em}\scalebox{0.72}{\textcolor{red}{(+#1)}}}

\title{AECG: Asymmetric Experience Consolidation and Governance In Multi-Agent Systems}

\author{
  \textbf{Ao Tian}$^{1}$,
  \textbf{Jialong Liu}$^{2}$,
  \textbf{Daqi Zheng}$^{3}$,
  \textbf{Xin Sun},
  \textbf{Mengting Li}$^{1}$, 
  \textbf{Zhizhao Xiao}$^{1}$,\\
  \textbf{Zijian Huang}$^{1}$,
  \textbf{Honglei Wang}$^{3}$,
  \textbf{Zijian Hei}$^{3}$, 
  \textbf{Yukun Yan}$^{4}$ \\
  $^{1}$Beihang University \quad
  $^{2}$Wuhan University \quad
  $^{3}$ModelBest \quad
  $^{4}$Tsinghua University
}

\begin{document}

\maketitle

\begin{abstract}
Large language model (LLM)-based multi-agent systems increasingly rely on memory to transform execution trajectories into reusable procedural knowledge. Yet repeated retrieval also makes memory errors persistent: memory pollution arises when outdated, weakly supported, or spuriously successful procedures become recurring components of future reasoning. Multi-agent execution introduces an additional structural risk. Scope collapse occurs when procedural knowledge escapes the coordination scope in which it was shown effective and is repeatedly reused at incompatible decision levels, allowing local errors to influence cascades of downstream decisions. Meanwhile, task-level failures provide ambiguous supervision because they rarely reveal which recalled knowledge was responsible. We introduce AECG, a framework for asymmetric experience consolidation and governance for multi-agent systems. AECG turns memory from static experience storage into a dynamic reliability-governance loop, preserving coordination scope
and using multi-scale, confidence-aware reliability to detect
degradation. It then combines degradation with downstream impact to
prioritize high-risk knowledge under a bounded review budget, applies targeted
interventions, and reactivates revised skills only after paired replay.
Across three multi-agent frameworks and four benchmarks, AECG achieves the best score in 11 of 12 framework--benchmark settings and improves over the strongest competing memory method by as much as 10.23 percentage points; removing scope preservation reduces accuracy by up to 16.89 points. AECG thereby reframes multi-agent memory from passive accumulation into auditable reliability governance.Code is
available at  https://github.com/fenhg297/AECG
\end{abstract}

\section{Introduction}

Large language model (LLM)-based multi-agent systems increasingly rely on
memory to transform execution trajectories into reusable procedural knowledge
\citep{sumers2023cognitive,wang2024survey,liu2024agentbench}. Prior work has
made substantial progress in experience representation, skill abstraction,
and memory retrieval, enabling agents to distill strategies from past
interactions and reuse them in subsequent tasks
\citep{park2023generative,shinn2023reflexion,zhao2024expel,wang2024agent,xu2026mem}.
Yet as memory evolves from trajectory storage into a library of persistently
reused skills, a deeper limitation emerges: existing approaches primarily
emphasize how knowledge is formed and recalled, leaving its reliability decay
under continued reuse comparatively underexplored. Correct retrieval does not
imply reliability, nor does it imply that the retrieved knowledge can be
safely generalized across different levels of collaboration.

Continued reuse makes this reliability gap increasingly consequential. An
isolated error typically affects only a local execution. However, once an
outdated, weakly supported, or spuriously successful procedure is consolidated
into long-term memory and repeatedly recalled, the error becomes a persistent
component of future reasoning. This gives rise to memory
pollution and sustained negative transfer across tasks
\citep{zhang2022survey,tang2026memory,wu2026bayesian,zhang2026useful}.

Multi-agent collaboration amplifies these risks. Its execution spans
multiple levels---including team planning, collaborative events, and
role-specific actions---whose interdependent decisions form a complex
execution structure
\citep{li2023camel,qian2024chatdev,hong2024metagpt,du2023improving,zhang2026g,han2025legomem}.
Reliable multi-agent memory governance must answer three questions.
First, where are the boundaries of valid reuse? Once a local procedure
escapes the coordination scope supported by its original evidence, it can
trigger scope collapse. Second, when has knowledge degraded?
Task-level failures are difficult to attribute to a particular planning,
coordination, or execution decision \citep{kong2026aegis}, while the inertia
of historical success can mask a recent reliability decline. Third,
how should intervention proceed under bounded computation? Skills
occupy different topological positions in the execution graph and therefore
have markedly different downstream impact; a system cannot review
every suspicious skill with an equal budget.

\begin{wrapfigure}{r}{0.52\textwidth}
\vspace{-0.8\baselineskip}
\centering
\includegraphics[width=\linewidth]{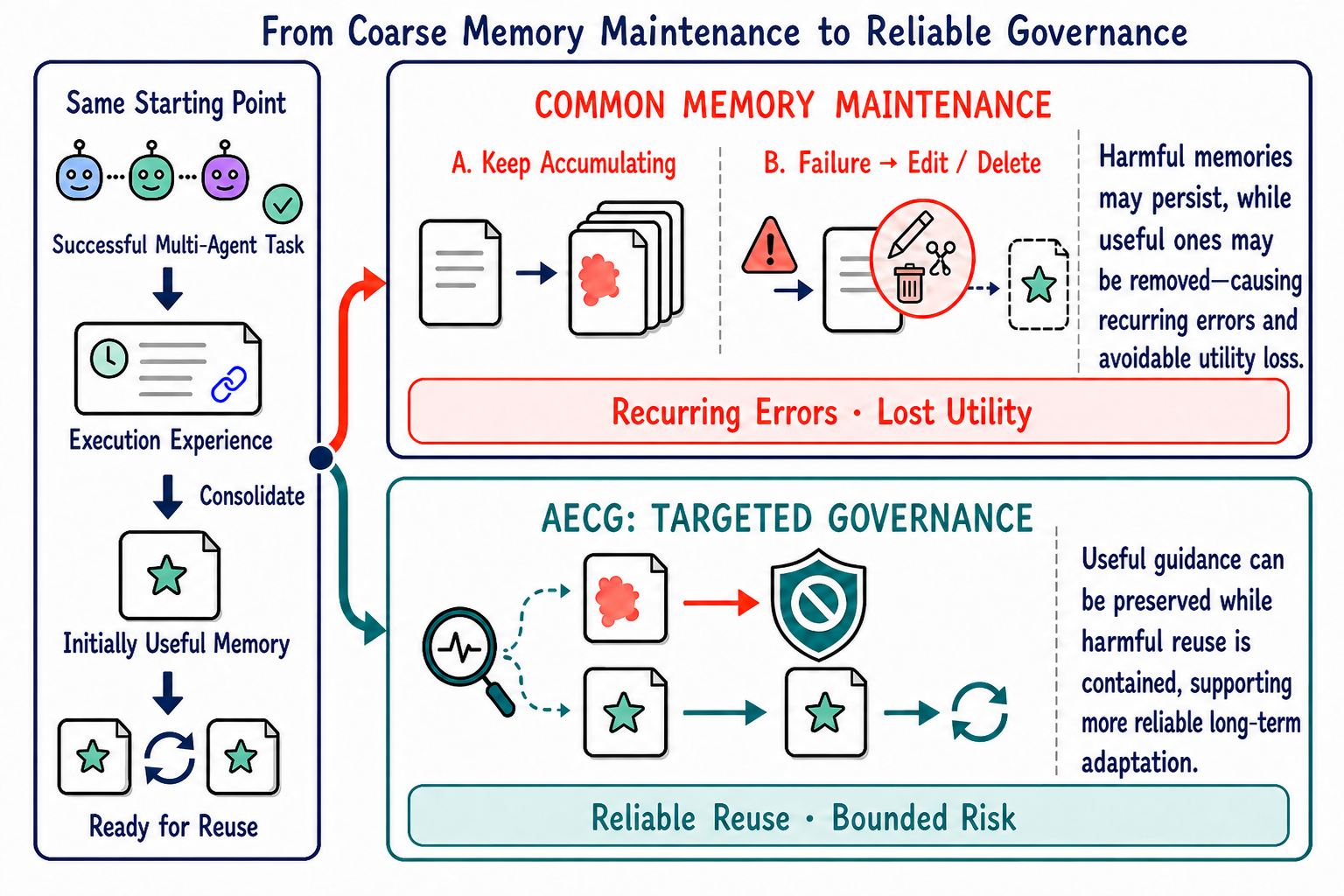}
\setlength{\abovecaptionskip}{1pt}
\caption{Conceptual comparison of memory maintenance: coarse accumulation or reactive revision risks persistent errors and utility loss, whereas AECG targets governance for reliable reuse.}
\label{fig:motivation-comparison}
\vspace{-1.2\baselineskip}
\end{wrapfigure}
Together, these questions require multi-agent memory to shift from passive
accumulation to continuous lifecycle governance, as illustrated in
Figure~\ref{fig:motivation-comparison}. To this end, we introduce
\textbf{AECG}, a framework for asymmetric experience consolidation and
governance. First, AECG preserves coordination scope throughout experience
storage, skill abstraction, and retrieval, ensuring knowledge is reused only
within evidence-supported boundaries. Second, it admits success and failure
asymmetrically: successful trajectories provide confidence-weighted positive
evidence, whereas a failure affects reliability only when its responsible scope
and associated skill are consistently attributed. Over the admitted evidence,
AECG maintains a cumulative slow belief and a fast belief that emphasizes
recent outcomes, revealing degradation otherwise masked by historical success.
Finally, it combines degradation with downstream exposure in the execution
graph to prioritize high-impact skills under a bounded review budget.

For each reviewed skill, AECG distinguishes failures caused by overly broad
applicability from those caused by a procedural defect and accordingly
proposes either context narrowing or procedural repair. A proposal creates a
new candidate version rather than overwriting the active skill. The candidate
is reactivated only if it mitigates the implicated behavior on failure
fixtures while preserving prior supported behavior on historical success
fixtures. By doing so, AECG closes the governance loop: it constrains where knowledge is reused, detects when it degrades, prioritizes intervention, and verifies whether a revised skill can safely return to service.

We evaluate AECG across three multi-agent frameworks and four benchmarks. It
achieves the best score in 11 of 12 settings, outperforming the strongest
memory baseline by up to 10.23 percentage points. Mechanism studies show
that removing scope preservation reduces accuracy by up to 16.89 points,
dual-timescale beliefs detect degradation faster, and downstream exposure
improves high-impact risk prioritization under a fixed review budget.

In summary, our main contributions are as follows:

\noindent\textbf{1.} We propose a scope-preserving, asymmetric experience
consolidation framework that maintains coordination boundaries during
knowledge formation and reuse while controlling negative-evidence admission
through conservative failure attribution.

\noindent\textbf{2.} We develop a dual-timescale reliability estimator that
compares historical and recent beliefs to identify skill degradation masked by
accumulated evidence of past success.

\noindent\textbf{3.} We develop topology-aware, budgeted governance that
combines reliability degradation with exposure and enables safe
skill updates through evidence-routed intervention and paired replay.

\section{Related Work}

\textbf{Memory and experience reuse in LLM agents.}
Early agent memory systems retrieve interaction histories, observations, or reflections to inform later decisions \citep{lewis2020retrieval,borgeaud2022improving,packer2023memgpt}. Generative Agents derives higher-level reflections from experience streams \citep{park2023generative}, while Reflexion and ExpeL retain feedback or transferable lessons from prior trajectories \citep{shinn2023reflexion,zhao2024expel}. More recent systems induce reusable workflows or continually organize memories into structured, revisable representations \citep{wang2024agent,xu2026mem}. This progression makes experience more actionable, but also turns generated abstractions---not only raw observations---into persistent inputs to future reasoning.

\textbf{Memory in multi-agent systems.}
Multi-agent memory additionally determines where experience resides, which agent consumes it, and how collaborative traces are represented. G-Memory organizes experience through hierarchical graphs, while LEGOMem separates full-task memories for orchestration from subtask memories for specialized agents \citep{zhang2026g,han2025legomem}. LLMA-Mem studies local, shared, and hybrid memory topologies, and ConMem coordinates useful strategies and failure cues in a relation graph \citep{wu2026scaling,tan2026conmem}. These designs highlight the importance of representation, placement, and inference-time coordination, but focus primarily on effective construction and retrieval rather than the longitudinal reliability of reused procedures or the propagation of their errors through collaboration.

\textbf{Procedural consolidation and skill evolution.}
Procedural memory compresses experience into reusable policies, workflows, or skills \citep{wei2022chain}. Voyager refines an executable skill library through environmental feedback \citep{wang2023voyager}, while MACLA constructs hierarchical procedures, maintains Beta reliability beliefs, and refines their conditions from successes and failures \citep{forouzandeh2025learning}. MSCE retains evidence links, applicability boundaries, verification rules, and reliability estimates during skill promotion \citep{tang2026memory}; Bayesian-Agent instead treats skills as evidence-bearing hypotheses and uses posterior beliefs to guide their evolution \citep{wu2026bayesian}. These methods move beyond simple accumulation, yet repeated LLM-based consolidation can still discard evidence or introduce spurious rules \citep{zhang2026useful}. AECG complements this line by jointly preserving coordination scope, admitting failures through conservative attribution, reassessing reliability at two timescales, and governing reuse via downstream exposure.

\section{Method}
\label{sec:method}

\subsection{Problem Formulation and Method Overview}
\label{sec:problem}

We consider a stream of multi-agent tasks $\{x_t\}_{t=1}^{T}$. The execution of each task is formalized as a directed execution graph $\mathcal{G}_t=(\mathcal{V}_t,\mathcal{E}_t)$, producing a corresponding trace $\tau_t$ and a terminal outcome $r_t\in\{0,1\}$. Each node $v\in\mathcal{V}_t$ represents a  planning, coordination, or execution decision, while the edges reflect the conditional dependencies between these decisions. To  capture the hierarchical nature of multi-agent collaboration, we map each node to a specific coordination scope:
\begin{equation}
s(v)\in\mathcal{S}_t
=
\{\textsc{Team}\}
\cup
\{\textsc{Event}(e)\}
\cup
\{\textsc{Agent}(e,a)\}.
\end{equation}
AECG maintains two scope-tagged objects: a procedural memory $m$ grounded in an execution trace and a reusable skill $k$ abstracted from supported procedural experience. Their scope constrains recall, evidence accumulation, abstraction, and comparison.

\begin{figure}[t]
\vspace{-0.7\baselineskip}
\centering
\includegraphics[width=\textwidth]{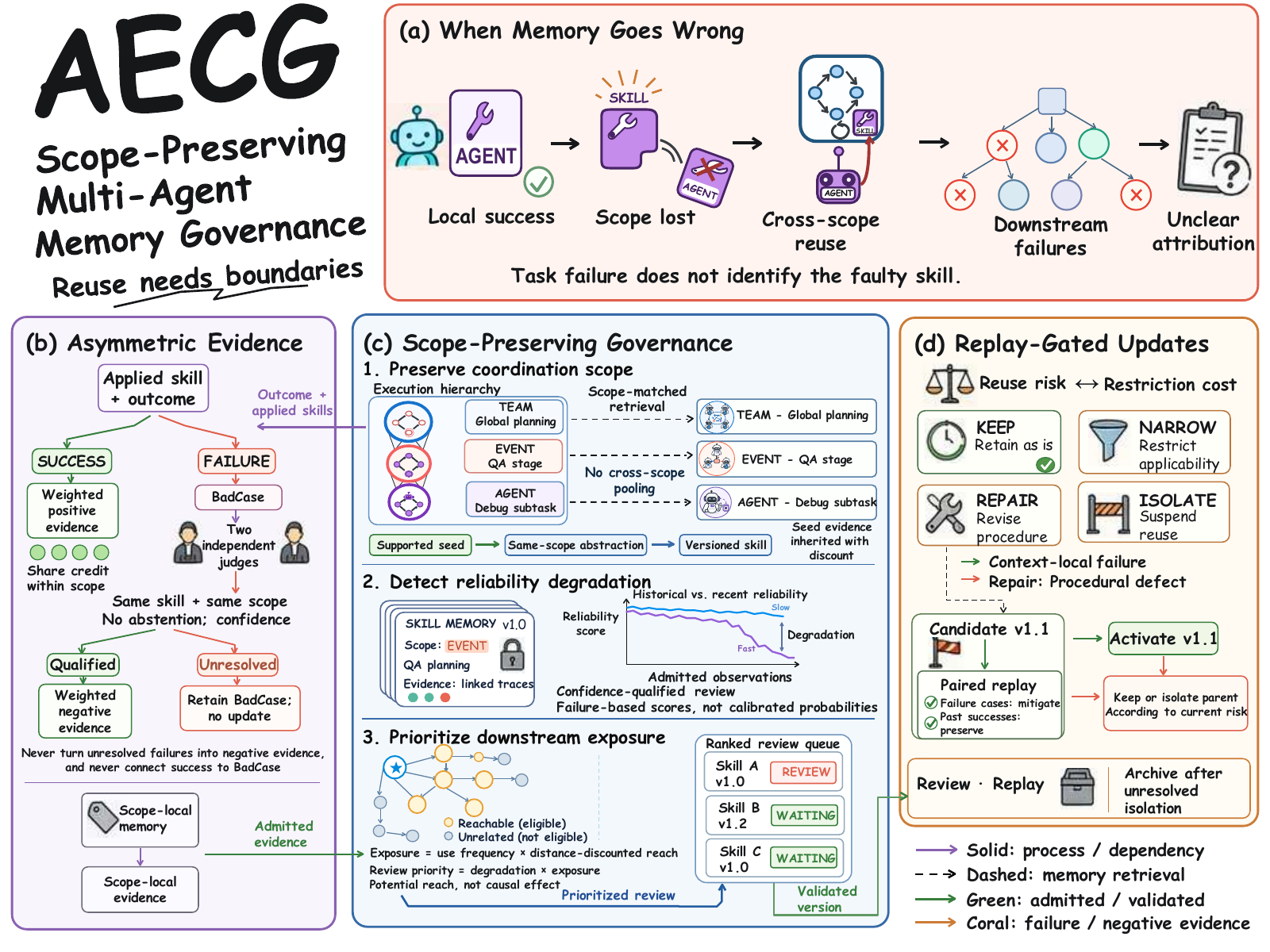}
\setlength{\abovecaptionskip}{1pt}
\caption{Overview of AECG. Agents retrieve scope-aligned procedural memories and skills. Positive evidence consolidates online, while ambiguous failures await qualified local feedback as BadCases. Scope-preserving promotion yields versioned skills, whose dual-timescale reliability and downstream exposure drive budgeted review and replay-gated governance.}
\label{fig:aegis-framework}
\vspace{-0.7\baselineskip}
\end{figure}

Persistent knowledge creates two opposing service risks. Reusing an unreliable skill can propagate errors through future executions, whereas restricting a useful skill can withhold guidance. AECG addresses this trade-off through a closed loop: it admits heterogeneous feedback as confidence-weighted evidence, maintains historical and recent Beta-style pseudo-count beliefs, and reviews confidence-qualified skills according to their reliability degradation and distance-discounted downstream exposure. Qualified evidence routes reviewed skill to context narrowing or procedural repair, and paired replay prevents unsafe revisions from activation. Figure~\ref{fig:aegis-framework} summarizes this lifecycle.

\subsection{Scope-Constrained Evidence Lifecycle}
\label{sec:consolidation}

To instantiate this governance loop, AECG must first control how execution
experience and feedback enter persistent memory and become reusable knowledge.
It does so through scope preservation, asymmetric evidence admission, and
evidence-grounded skill formation.
\subsubsection{Scope-preserving memory objects}

Team-level objects support global planning, event-level objects support a collaborative stage, and agent-level objects support role-specific execution while retaining their source identity. Evidence is not pooled across these levels. When procedural memory $m_{seed}$ is abstracted into skill $k$, AECG enforces $s(k) = s(m_{seed})$. This constraint is applied during recall and comparison, preventing local experience from becoming system-wide knowledge and causing scope collapse.
\subsubsection{Asymmetric evidence admission}
\label{sec:evidence-admission}

Scope preservation determines where evidence may be applied; it does not,
however, determine whether an observed outcome provides reliable evidence.
Task outcomes are inherently asymmetric in this respect: success provides
confidence-weighted positive evidence for skills applied in a successful
trajectory, whereas task-level failure alone does not identify which
persistent object deserves negative evidence. For the $u$-th use of skill
$k$, let $y_{k,u}\in\{1,0,\emptyset\}$ denote admitted positive, admitted
negative, or unresolved evidence, and let $c_{k,u}\in[0,1]$ be its confidence
weight. Given task outcome $r_t$, AECG routes evidence as
\begin{equation}
\label{eq:asymmetric-routing}
(y_{k,u},c_{k,u})=
\begin{cases}
(1,c^{+}_{k,u}), &
r_t=1\ \text{and } k\text{ was applied},\\
(\emptyset,0), &
r_t=0\ \text{and no qualified attribution to }k\text{ is available},\\
(0,c^{\mathrm{att}}_{k,u}), &
r_t=0\ \text{and }k\text{ passes the attribution gate}.
\end{cases}
\end{equation}

Qualified attribution uses the failed-task execution record, including the
applied skills, scopes, and execution nodes. Two LLM
judges may abstain and can select only applied skills. Negative evidence is
admitted only if both identify the same skill and scope, neither abstains,
and their mean confidence meets a predefined attribution threshold; otherwise the observation
remains unresolved. Thus, attribution serves as conservative evidence
admission rather than ground-truth causal identification.  Failed trajectories remain available as
BadCases. Appendix~\ref{app:evidence-attribution} details the protocol.

This admission rule induces, for each skill, the set of observations that may
affect its reliability:
\begin{equation}
\mathcal A_k
=
\{u\mid y_{k,u}\in\{0,1\}\}.
\label{eq:admitted-set}
\end{equation}
Only observations in $\mathcal A_k$ update the reliability beliefs.
\subsubsection{Skill formation and evidence inheritance}

Having established which observations count as evidence, AECG next determines
when a same-scope procedural memory has accumulated sufficient support to
become a reusable skill and how its evidence should be inherited after
abstraction. Repeatedly supported procedural memories become candidates for
skill formation through a within-scope readiness rule. Its implementation is deferred to Appendix~\ref{app:evidence-promotion}. An
LLM abstracts the selected seed into a reusable same-scope skill. Nearby
memories may clarify applicability, but only the seed supplies initial
evidence.

Let $n_{\mathrm{seed}}^{+}$ and $n_{\mathrm{seed}}^{-}$ denote the numbers of admitted positive and negative observations, $n_{\mathrm{seed}}=n_{\mathrm{seed}}^{+}+n_{\mathrm{seed}}^{-}$, and $\widehat p_{\mathrm{seed}}=n_{\mathrm{seed}}^{+}/n_{\mathrm{seed}}$. For $q\in\{S,F\}$, define
\[
\mathbf b_{k,u}^{q}=
\begin{bmatrix}\alpha_{k,u}^{q}\\ \beta_{k,u}^{q}\end{bmatrix},
\qquad
\mathbf b_0=
\begin{bmatrix}\alpha_0\\ \beta_0\end{bmatrix},
\qquad
\mathbf e(p)=
\begin{bmatrix}p\\1-p\end{bmatrix}.
\]
With initial  pseudo-counts $\alpha_0=\beta_0=1$, AECG transfers a fraction $\kappa\in(0,1]$ of the seed evidence:
\begin{equation}
\label{eq:evidence-transfer}
n_{k,0}^{S}=\kappa n_{\mathrm{seed}}, \qquad
n_{k,0}^{F}=\min\{W,\kappa n_{\mathrm{seed}}\}, \qquad
\mathbf b_{k,0}^{q}=\mathbf b_0+n_{k,0}^{q}\mathbf e(\widehat p_{\mathrm{seed}}) 
\end{equation}
The slow state retains all discounted seed evidence, whereas the fast state caps inherited mass at window $W$. The coefficient $\kappa$ controls how much seed evidence is inherited by the abstracted skill, limiting the influence of evidence accumulated before abstraction.
\subsection{Dual-Timescale Reliability Estimation}
\label{sec:reliability}

Once a skill is formed with qualified evidence inherited from its seed memory,
AECG must translate admitted outcomes into evolving reliability estimates.
To separate historical support from recent effectiveness, each active skill
version maintains two Beta-style pseudo-count beliefs: one slow and
cumulative, and the other fast and discounted. We use these Beta distributions
for scoring, rather than as posteriors under a full generative Bayesian model:
\begin{equation}
\theta_{k,u}^{S}\sim\operatorname{Beta}(\alpha_{k,u}^{S},\beta_{k,u}^{S}),
\qquad
\theta_{k,u}^{F}\sim\operatorname{Beta}(\alpha_{k,u}^{F},\beta_{k,u}^{F}).
\end{equation}

The slow belief provides a historical baseline, whereas the fast belief represents recent effectiveness. For $u\in\mathcal A_k$, let $\operatorname{prev}_k(u)$ be the preceding index in $\mathcal A_k$, with $\mathbf b_{k,\operatorname{prev}_k(u)}^{q}:=\mathbf b_{k,0}^{q}$ if none exists. Setting $\lambda_S=1$ and $\lambda_F=\lambda<1$, each admitted observation updates both beliefs as
\begin{equation}
\mathbf b_{k,u}^{q}
=
\mathbf b_0+\lambda_q\bigl(\mathbf b_{k,\operatorname{prev}_k(u)}^{q}-\mathbf b_0\bigr)
+c_{k,u}\mathbf e(y_{k,u}),
\quad u\in\mathcal A_k.
\label{eq:belief-update}
\end{equation}
The slow belief accumulates admitted evidence, while the fast belief discounts old evidence toward the initial pseudo-count state \citep{kim2026discounted}. 

The resulting beliefs support complementary assessments of current risk and
change over time. Their means and the current failure-risk score are
\begin{equation}
    \mu_k^q = \mathbb{E}[\theta_k^q] = \frac{\alpha_k^q}{\alpha_k^q + \beta_k^q}, \quad R_k^{\text{fail}} = 1 - \mu_k^F = \frac{\beta_k^F}{\alpha_k^F + \beta_k^F}.
\end{equation}
Here $R_k^{\mathrm{fail}}$ scores the risk of continuing to use the current version. At the belief-mean level, recent degradation relative to the historical baseline is
\begin{equation}
\label{eq:mean-drift}
D_k
=
\left[\mu_k^{S}-\mu_k^{F}\right]_+,
\qquad [x]_+=\max(0,x).
\end{equation}
Thus, the dual beliefs distinguish a skill that has always been unreliable from one whose recent effectiveness has declined relative to its own history. In implementation, AECG replaces this mean-only signal with a Beta-style belief-sampling estimate $R_k^{\mathrm{drift}}$ that discounts discrepancy attributable to sampling noise. The three quantities serve distinct roles: $R_k^{\mathrm{fail}}$ measures current failure tendency, $D_k$ provides an interpretable mean-level measure of recent degradation, and $R_k^{\mathrm{drift}}$ is the uncertainty-aware degradation score used for governance. Accordingly, $R_k^{\mathrm{fail}}$ is diagnostic, whereas periodic review uses confidence-gated $R_k^{\mathrm{drift}}$ and ranks eligible skills jointly with downstream exposure. The estimator and the uncertainty statistic used for audit are provided in Appendix~\ref{app:reliability-details}.
\subsection{Risk- and Exposure-Aware Memory Governance}
\label{sec:governance}

The dual-timescale estimator provides confidence-aware evidence of skill
degradation, but degradation alone does not determine how limited review
capacity should be allocated. Because reviewed skills differ in their
potential downstream impact, AECG combines reliability degradation with
structural exposure before committing intervention resources. Governance then
proceeds in three steps: budgeted triage, evidence-routed intervention, and
replay-gated activation.

\subsubsection{Exposure-aware budgeted triage}

Let $d_{k,u}\in\mathcal V_t$ be the node at which use $u$ selected skill $k$, let $\operatorname{Desc}_t(d):=\{v\in\mathcal V_t\setminus\{d\}:d\leadsto v\}$ denote its unique reachable descendants, and let $\operatorname{dist}_t(d_{k,u},v)$ be the shortest directed-path distance to a downstream node $v$. Its observed propagation exposure is
\begin{align}
\label{eq:observed-exposure}
e_{k,u}^{\mathrm{obs}}
&=
\sum_{v\in\operatorname{Desc}_t(d_{k,u})}
\gamma^{\operatorname{dist}_t(d_{k,u},v)-1},
\qquad 0<\gamma\leq 1,\\
\label{eq:predicted-exposure}
\bar e_k
&=
\frac{1}{M_k}\sum_{u\in\mathcal U_k^{\mathrm{rec}}}
e_{k,u}^{\mathrm{obs}},
\qquad
E_k=f_k^{\mathcal H}\bar e_k.
\end{align}
Thus, distance discounts remote descendants, while $E_k$ combines recent
mean exposure over $M_k=|\mathcal U_k^{\mathrm{rec}}|$ uses with the preceding
horizon's selected-use count $f_k^{\mathcal H}$. This is potential propagation
exposure, not a causal effect estimate.

Exposure alone does not indicate that a skill has degraded; it quantifies only
the potential reach of its continued reuse. AECG therefore applies exposure
only after reliability evidence has qualified a skill for review. Specifically,
the belief-sampling confidence score $C_k^{\mathrm{deg}}$ admits a skill to
periodic review only when $C_k^{\mathrm{deg}}\geq\eta_{\mathrm{deg}}$ and
$R_k^{\mathrm{drift}}>0$. Eligible skills are ranked by
\begin{equation}
\label{eq:priority}
H_k:=E_kR_k^{\mathrm{drift}},
\end{equation}
and up to the top $B_{\mathrm{review}}$ enter the  review set. Qualified
local failures use an event-driven budget. Confidence estimation,
cold-start handling, and budget rules are detailed in
Appendix~\ref{app:governance-details}.

\subsubsection{Evidence-routed and replay-gated intervention}

Once a skill enters review---through periodic exposure-aware triage or
qualified event-driven failure handling---the question is how it should
be treated. AECG uses qualified failure evidence to determine the intervention
type. Context-localized failures yield \textsc{Narrow}, which restricts
applicability without changing the procedure; cross-context defects
yield \textsc{Repair}, which revises the procedure while preserving scope and
intent. If neither condition is supported, no candidate is generated.

Selecting an intervention type is not sufficient for safe deployment: a
proposed correction may resolve the observed failure while damaging previously
supported behavior. AECG therefore creates a versioned candidate rather than
modifying the active skill in place and evaluates it through paired replay.
Failure fixtures test whether the implicated behavior is mitigated, while
disjoint regression fixtures from prior successful same-scope uses test whether
supported behavior is preserved. Let
$\widehat r_{k,z}^{\mathrm{fix}}$ and $\widehat r_{k,z}^{\mathrm{reg}}$ denote
candidate success on the two sets, and let
$\widehat r_{k,\mathrm{parent}}^{\mathrm{reg}}$ denote the active parent's
regression success. The replay gate is
\begin{equation}
\label{eq:replay-gate}
\operatorname{Gate}(z,k)
:=
\mathbf 1\{\widehat r_{k,z}^{\mathrm{fix}}=1\}
\cdot
\mathbf 1\{
\widehat r_{k,z}^{\mathrm{reg}}
\geq
\widehat r_{k,\mathrm{parent}}^{\mathrm{reg}}-\epsilon_{\mathrm{reg}}
\},
\qquad
z\in\{\textsc{Narrow},\textsc{Repair}\},
\end{equation}
A candidate is activated only if it mitigates the implicated behavior and
preserves prior supported behavior within tolerance $\epsilon_{\mathrm{reg}}$.
Otherwise, fast-belief confidence routes the skill to continued monitoring or
isolation. Action-specific replay and fallback rules are given in
Appendix~\ref{app:governance-details}.
\section{Experiments}
\label{sec:experiments}

Our experiments test four hypotheses derived from the lifecycle-governance
perspective. \textbf{RQ1} asks whether AECG improves end-task performance
across frameworks and domains; \textbf{RQ2} asks whether its governance
components are necessary and, whether scope preservation
prevents cross-level misuse from inducing scope collapse; \textbf{RQ3} asks
whether dual-timescale beliefs overcome the inertia of historical success to
detect skill degradation promptly; and \textbf{RQ4} asks whether downstream
exposure improves the allocation of limited review capacity toward
consequential degraded skills.

\subsection{Experimental Setup}
\label{sec:experimental-setup}

We evaluate AECG on three MAS frameworks---AutoGen
\citep{wu2023autogen}, MacNet \citep{qian2025scaling}, and AgentNet
\citep{yang2026agentnet}---and four benchmarks spanning planning, fact
verification, interactive reasoning, and code generation: PDDL
\citep{ma2024agentboard}, FEVER \citep{thorne2018fever}, ScienceWorld
\citep{wang2022scienceworld}, and MBPP-Plus \citep{liu2023your}. We compare
against No-Memory, MacNet-M, MemoryBank \citep{zhong2024memorybank}, ConMem
\citep{tan2026conmem}, and G-Memory \citep{zhang2026g}. All methods share
tasks, prompts, tools, execution limits, and evaluators; execution uses
\texttt{gpt-4.1-mini}, memory-layer calls use \texttt{gpt-4o-mini}, and results
average three independently seeded runs.  Mechanism tests use AgentNet--ScienceWorld with a
shared post-warm-up memory snapshot and matched conditions. Appendix~\ref{app:experimental-details}
provides full setup details.

\subsection{RQ1: Does AECG Improve End-Task Performance?}
\label{sec:main-results}

\begin{table}[t]

\vspace{-0.5em}
\caption{Performance comparison across frameworks and memory methods, averaged over 3 seeds. Bold indicates the best result, with parentheses showing the absolute percentage-point margin over the runner-up. AECG rows are shaded light green.}
\label{tab:main-results}
\vspace{0.3em}
\centering
\small
\setlength{\aboverulesep}{0pt}
\setlength{\belowrulesep}{0pt}
\setlength{\tabcolsep}{0pt}
\renewcommand{\arraystretch}{1.15}
\begin{tabular}{
  p{0.12\linewidth}<{\centering} 
  p{0.14\linewidth} 
  p{0.15\linewidth}<{\centering} 
  p{0.14\linewidth}<{\centering} 
  p{0.15\linewidth}<{\centering} 
  p{0.16\linewidth}<{\centering} 
  p{0.14\linewidth}<{\centering}
}
\toprule
\rowcolor{gray!15}
\textbf{MAS} & \textbf{Memory} & \textbf{PDDL} & \textbf{FEVER} & \textbf{SciWorld} & \textbf{MBPP-Plus} & \textbf{Avg.} \\
\midrule
\multirow{6}{*}{AutoGen}
& No-Memory  & 36.52 & 71.42 & 58.62 & 74.59 & 63.43 \\
& MacNet-M   & 31.78 & 63.91 & 57.78 & 76.23 & 60.97 \\
& MemoryBank & 40.12 & 68.17 & 56.70 & 75.41 & 63.01 \\
& ConMem     & 40.23 & 69.84 & 55.56 & 73.77 & 62.66 \\
& G-Memory   & 38.33 & 67.61 & 60.02 & 77.05 & 63.87 \\
\rowcolor{green!10} \cellcolor{white}
& AECG       & \textbf{46.67}\gain{6.44} & \textbf{74.49}\gain{3.07} & \textbf{63.35}\gain{3.33} & \textbf{82.78}\gain{5.73} & \textbf{69.79}\gain{5.92} \\
\midrule
\multirow{6}{*}{MacNet}
& No-Memory  & 45.31 & 61.22 & 47.73 & 75.41 & 59.95 \\
& MacNet-M   & 48.44 & 67.35 & 60.23 & 77.87 & 65.86 \\
& MemoryBank & 48.44 & 68.37 & 51.14 & 71.05 & 61.74 \\
& ConMem     & \textbf{54.12} & 70.41 & 61.36 & 73.77 & 66.57 \\
& G-Memory   & 49.21 & 69.39 & 44.32 & 71.31 & 60.62 \\
\rowcolor{green!10} \cellcolor{white}
& AECG       & 51.56 & \textbf{71.43}\gain{1.02} & \textbf{71.59}\gain{10.23} & \textbf{79.51}\gain{1.64} & \textbf{70.70}\gain{4.13} \\
\midrule
\multirow{6}{*}{AgentNet}
& No-Memory  & 45.31 & 59.18 & 55.68 & 68.85 & 59.14 \\
& MacNet-M   & 51.56 & 66.33 & 55.68 & 72.13 & 63.17 \\
& MemoryBank & 53.13 & 68.37 & 62.50 & 69.67 & 64.79 \\
& ConMem     & 50.06 & 69.39 & 59.09 & 71.31 & 64.26 \\
& G-Memory   & 48.44 & 69.22 & 42.05 & 63.11 & 57.21 \\
\rowcolor{green!10} \cellcolor{white}
& AECG       & \textbf{54.69}\gain{1.56} & \textbf{73.47}\gain{4.08} & \textbf{68.18}\gain{5.68} & \textbf{77.87}\gain{5.74} & \textbf{70.43}\gain{5.64} \\
\bottomrule
\end{tabular}
\vspace{-1.0em}
\end{table}
Table~\ref{tab:main-results} shows that AECG achieves the best score in 11 of
12 framework--benchmark settings and ranks first in average performance for three frameworks. Its margin over the strongest memory method
is 5.92 on AutoGen, 4.13 on MacNet, and 5.64 on AgentNet. Unlike
methods that win only under a particular topology, AECG remains
strong across conversational, graph-orchestrated, and decentralized systems.
This consistency is the primary evidence for the paper's overall claim:
reusable experience benefits from lifecycle governance rather than
accumulation and retrieval alone.

The framework-wise pattern also clarifies where governance matters. AECG
improves over No-Memory by 6.36 points on AutoGen, 10.75 on MacNet, and 11.29
on AgentNet in average score. The larger gains on graph-structured MacNet and
AgentNet are consistent with the motivation that a reused procedure can
affect more dependent decisions when collaboration exposes explicit routing
or dependency structure. Importantly, AECG also improves AutoGen, showing that
its benefit is not restricted to frameworks with an explicit execution DAG.

Across benchmarks, the largest margin over the runner-up is 10.23 points on
MacNet--ScienceWorld, while MBPP-Plus improves consistently across all three
frameworks. Both settings involve repeated use of intermediate procedures and
therefore stress stale or over-generalized skills. PDDL is the most
competitive case: ConMem exceeds AECG by 2.56 points on MacNet, the sole
exception among 12 settings. Thus, RQ1 is answered positively at the aggregate
level, while also showing that no single memory organization dominates every
framework--task pairing. Appendix~\ref{app:main-results-extra} provides the
full heatmap and additional benchmark-wise analysis.
\subsection{RQ2: Are the Governance Components Necessary, and Does Scope Preservation Prevent Scope Collapse?}
\label{sec:ablation}

Table~\ref{tab:ablation} first tests whether the three contributions are
functionally necessary rather than merely descriptive additions. Removing
scope preservation causes the largest degradation: success drops by 16.89
points on ScienceWorld and 8.34 on MBPP-Plus, establishing scope preservation
as the most consequential component in this ablation.

\begin{wraptable}{r}{0.52\textwidth}
\vspace{-1.7\baselineskip}
\caption{AgentNet ablations. Parentheses show drops from AECG.}
\label{tab:ablation}
\centering
\scriptsize
\setlength{\aboverulesep}{0pt}
\setlength{\belowrulesep}{0pt}
\setlength{\tabcolsep}{0pt}
\renewcommand{\arraystretch}{1.1}
\begin{tabular}{p{0.34\linewidth} p{0.33\linewidth}<{\centering} p{0.33\linewidth}<{\centering}}
\toprule
\rowcolor{gray!15}
\textbf{Variant} & \textbf{SciWorld} & \textbf{MBPP+} \\
\midrule
\textbf{Full AECG} & \textbf{68.18} & \textbf{77.87} \\
\midrule
w/o Scope     & 51.29 {\color{red}(-16.89)} & 69.53 {\color{red}(-8.34)} \\
w/o Dual-Time & 63.42 {\color{red}(-4.76)}  & 71.61 {\color{red}(-6.26)} \\
w/o Exposure  & 64.82 {\color{red}(-3.36)}  & 75.29 {\color{red}(-2.58)} \\
\bottomrule
\end{tabular}
\vspace{-1.0\baselineskip}
\end{wraptable}
The other ablations connect the aggregate result to Contributions 2 and 3.
Without the fast belief, performance drops by 4.76 on ScienceWorld and
6.26 on MBPP-Plus, indicating that cumulative evidence alone is too inert
when recent effectiveness changes. Removing exposure from the review priority
also degrades both domains despite retaining the same confidence gate and
review budget. Table~\ref{tab:ablation} therefore establishes necessity; the
remaining RQs test whether the intended mechanisms explain these losses.

\begin{wraptable}{r}{0.52\textwidth}
\vspace{-1.95\baselineskip}
\caption{Scope preservation on AgentNet--ScienceWorld.}
\label{tab:scope-mechanism}
\centering
\scriptsize
\setlength{\aboverulesep}{0pt}
\setlength{\belowrulesep}{0pt}
\setlength{\tabcolsep}{0pt}
\renewcommand{\arraystretch}{1.1}
\begin{tabular}{p{0.28\linewidth} p{0.24\linewidth}<{\centering} p{0.24\linewidth}<{\centering} p{0.24\linewidth}<{\centering}}
\toprule
\rowcolor{gray!15}
\textbf{Method} & \textbf{Success} $\uparrow$ & \textbf{Cross-scope} $\downarrow$ & \textbf{Exposure} $\downarrow$ \\
\midrule
\textbf{Hard-scope} & \textbf{68.18} & \textbf{0.0} & -- \\
Flat-scope          & 51.29          & 36.5         & 14.8 \\
\bottomrule
\end{tabular}
\vspace{-0.8\baselineskip}
\end{wraptable}
Flattening scope causes 36.5\% of selected uses to cross their supported
coordination scope and reduces success by 16.89
(Table~\ref{tab:scope-mechanism}). These cross-scope uses have average
failed-task exposure of 14.8, indicating that unsupported reuse can
reach multiple downstream decisions. Hard matching eliminates cross-scope
reuse. Because exposure is a structural reachability measure, this
result associates scope collapse with lower success and broader
reach without claiming every descendant was causally changed.

Crucially, the Flat-scope variant reaches only 51.29 on ScienceWorld, falling
below the No-Memory baseline of 55.68. Persistent memory therefore becomes
counterproductive when knowledge is reused beyond the coordination scope
supported by its evidence. Together with the observed 36.5\% cross-scope use
rate and broader potential exposure, this result supports the central
hypothesis behind RQ2: all three governance components contribute, and scope
preservation is a necessary safeguard against negative transfer from scope
collapse.

\subsection{RQ3: Can Dual-Timescale Reliability Overcome Historical Inertia?}
\label{sec:mechanistic-analysis}

We inject controlled skill corruptions to test whether the fast belief makes
recent contradictory evidence actionable.
\begin{figure}[H]
\vspace{-0.7\baselineskip}
\centering
\includegraphics[width=0.9\linewidth,trim=0 0 0 28,clip]{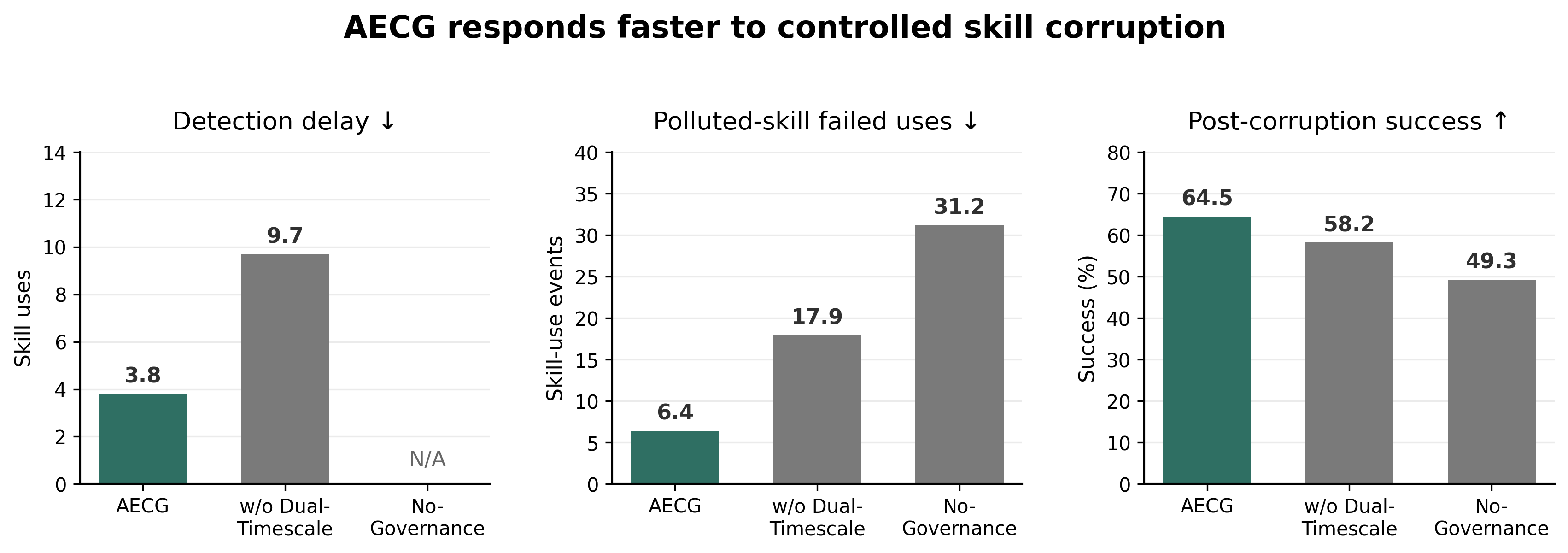}
\setlength{\abovecaptionskip}{2pt}
\caption{Detection and recovery after controlled skill corruption.}
\label{fig:controlled-corruption}
\vspace{-0.7\baselineskip}
\end{figure}
Figure~\ref{fig:controlled-corruption} separates rapid detection from the
general benefit of having a governance loop. AECG reviews a corrupted skill
after 3.8 uses on average, compared with 9.7 for the cumulative-only variant,
a 60.8\% reduction in delay. Earlier review reduces polluted-skill failed uses
from 17.9 to 6.4 and improves post-corruption success from 58.2\% to 64.5\%.

No-Governance provides a second reference point. It continues to retrieve the
corrupted skills but never reviews them, accumulating 31.2 polluted-skill
failed uses and reaching only 49.3\% success. Thus, detecting drift earlier is
useful because it triggers intervention before pollution persists through the
remaining task stream. Because AECG and the cumulative-only variant receive
the same historical evidence, the reduction in detection delay isolates the
role of the fast belief: it prevents recent contradictory outcomes from being
overwhelmed by accumulated success. RQ3 therefore confirms that the
dual-timescale design counteracts historical inertia, allowing degraded skills
to be reviewed earlier and limiting their persistence as memory pollution.

\subsection{RQ4: Does Downstream Exposure Improve Triage Under a Fixed Budget?}

We compare exposure-aware, risk-only, and random selection under the same
confidence gate, corruption set, and review budget. Qualified candidates
outnumber available reviews, so the experiment isolates prioritization rather
than degradation detection or intervention quality.
\begin{figure}[H]
\vspace{-0.7\baselineskip}
\centering
\includegraphics[width=0.9\linewidth,trim=0 0 0 28,clip]{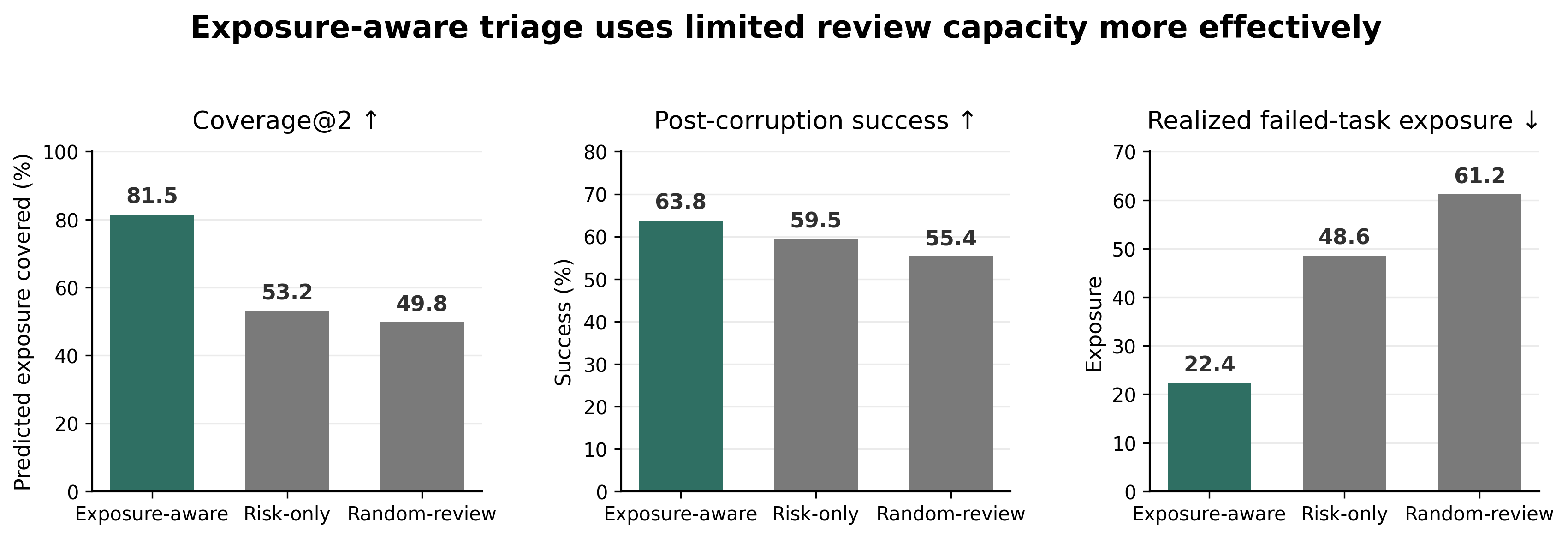}
\setlength{\abovecaptionskip}{2pt}
\caption{Triage under the same confidence gate and fixed review budget.}
\label{fig:budgeted-triage}
\vspace{-0.7\baselineskip}
\end{figure}
Exposure-aware ranking covers 81.5\% of polluted-skill predicted exposure in
the first two reviews, compared with 53.2\% for risk-only and 49.8\% for
random review (Figure~\ref{fig:budgeted-triage}). Coverage@2 is only a ranking
diagnostic; the primary result is post-corruption success, where exposure-aware
triage improves by 4.3 points over risk-only and 8.4 over random review under
the same budget.

The realized traces support the same conclusion without reusing the predicted
exposure that defines the ranking. Exposure-aware triage reduces failed-task
exposure to 22.4, versus 48.6 for risk-only and 61.2 for random review. Because
all methods use the same confidence gate, corruption set, intervention rule,
and review budget, the performance gain isolates the value of prioritization
rather than differences in degradation detection or repair quality. RQ4
supports the governance hypothesis: in a multi-agent execution
graph, downstream exposure provides an effective structural signal for
directing bounded review capacity toward degraded skills with greater
potential propagation impact.
\section{Conclusion}

We introduce AECG, shifting multi-agent procedural memory from passive accumulation toward lifecycle governance. AECG preserves where knowledge can be reused, detects when it degrades through asymmetric evidence and dual-timescale beliefs, and prioritizes which skills to review using downstream exposure. Evidence-routed intervention and paired replay enable safe skill updates. Across three multi-agent frameworks and four benchmarks, AECG achieves the best performance in 11 of 12 settings. Mechanism studies show that scope preservation mitigates cross-level negative transfer, dual-timescale beliefs overcome historical inertia, and downstream exposure improves review allocation. These results demonstrate that reliable memory requires continuous governance of reuse boundaries, reliability, and potential impact. AECG currently depends on the calibration of localized feedback and treats graph-based exposure as a structural proxy rather than a true causal estimate. Future work will model uncertain responsibility jointly with reliability and replace reachability with intervention-grounded causal influence.
\section*{AI Use Statement}
Generative AI tools were used to assist with English translation and language
polishing, software implementation, and the preparation of selected figures.
All AI-assisted text was reviewed and revised by the authors. AI-assisted code
was inspected and tested, and AI-assisted figures were checked against the
underlying methods and results. The authors take full responsibility for the
final content of this work, including all text, code, and figures produced with
AI assistance.

\bibliography{iclr2027_conference}
\bibliographystyle{iclr2027_conference}
\clearpage
\appendix

\section*{Appendices}
Within this supplementary material, we elaborate on the following aspects:

\begin{itemize}[apptoc]
    \item Appendix~\ref{app:algorithm}: Algorithm
    \item Appendix~\ref{app:method-details}: Method Implementation Details
    \begin{itemize}[apptoc]
        \item Appendix~\ref{app:evidence-promotion}: Evidence Assignment and Skill Promotion
        \item Appendix~\ref{app:reliability-details}: Beta-Style Belief Drift and Uncertainty (Proposition~1)
        \item Appendix~\ref{app:governance-details}: Triage and Action-Specific Validation
        \item Appendix~\ref{app:hyperparameters}: Hyperparameter Selection
        \item Appendix~\ref{app:evidence-attribution}: LLM-based Failure Attribution
    \end{itemize}
    \item Appendix~\ref{app:experimental-details}: Experimental Details
    \begin{itemize}[apptoc]
        \item Appendix~\ref{app:benchmarks-baselines}: Benchmarks, Baselines, Frameworks, and Models
        \item Appendix~\ref{app:main-results-extra}: Additional Main-Result Analysis
        \item Appendix~\ref{app:mechanism-protocols}: Mechanism-Test Protocols
    \end{itemize}
    \item Appendix~\ref{app:prompt}: Prompt Templates
    \begin{itemize}[apptoc]
        \item Appendix~\ref{app:prompt-skill-abstraction}: Skill Abstraction
        \item Appendix~\ref{app:prompt-failure-attribution}: Conservative Failure Attribution
        \item Appendix~\ref{app:prompt-context-narrowing}: Context Narrowing
        \item Appendix~\ref{app:prompt-procedural-repair}: Procedural Repair
        \item Appendix~\ref{app:prompt-replay-contract}: Replay Contract
    \end{itemize}
\end{itemize}

\section{Algorithm}
\label{app:algorithm}

\begin{algorithm}[H]
\caption{AECG: Asymmetric Experience Consolidation and Governance}
\small
\begin{algorithmic}[1]
\REQUIRE $\{x_t\}_{t=1}^{T}$; memory store $\mathcal M$;
skill registry $\mathcal K$; configured parameters and budgets
\ENSURE Updated $(\mathcal M,\mathcal K)$
\FOR{$t=1,\ldots,T$}
    \STATE \textbf{// Scope-preserving execution}
    \STATE Execute $x_t$ with applicable, scope-matched recall
    at each $v$; obtain $(\mathcal G_t,\tau_t,r_t)$

    \STATE \textbf{// Asymmetric consolidation}
    \IF{$r_t=0$}
        \STATE Retain $\tau_t$ as a BadCase;
        compute $G_{k,t}$ by Eq.~\ref{eq:attribution-gate}
    \ENDIF
    \STATE $(y_{k,u},c_{k,u})\leftarrow$
    Eq.~\ref{eq:asymmetric-routing}
    \STATE Update $\mathcal M$ from $\tau_t$ and admitted
    scope-local evidence
    \FOR{each newly admitted $u\in\mathcal A_k$}
        \STATE $\mathbf b_{k,u}^{q}\leftarrow
        \mathbf b_0+\lambda_q
        (\mathbf b_{k,\operatorname{prev}_k(u)}^{q}-\mathbf b_0)
        +c_{k,u}\mathbf e(y_{k,u}),\quad q\in\{S,F\}$
    \ENDFOR
    \FOR{each $m_{\mathrm{seed}}$ passing Eq.~\ref{eq:promotion-gate}}
        \STATE Abstract $k$ with $s(k)=s(m_{\mathrm{seed}})$
        \STATE Register $k$ using Eq.~\ref{eq:evidence-transfer};
        link equivalents without duplicate transfer
    \ENDFOR

    \STATE \textbf{// Exposure-aware review}
    \STATE $\mathcal Q^*\leftarrow\emptyset$
    \IF{periodic review is due}
        \STATE $(R_k^{\mathrm{drift}},C_k^{\mathrm{deg}})
        \leftarrow$ Eqs.~\ref{eq:drift}, \ref{eq:degradation-confidence}
        \STATE $\mathcal C_{\mathrm{deg}}\leftarrow
        \{k\in\mathcal K:\ k\text{ active},\
        C_k^{\mathrm{deg}}\geq\eta_{\mathrm{deg}},\
        R_k^{\mathrm{drift}}>0\}$
        \STATE $E_k\leftarrow$ Eq.~\ref{eq:predicted-exposure},
        $\quad H_k\leftarrow E_kR_k^{\mathrm{drift}}$
        \STATE $\displaystyle
        \mathcal Q^*\leftarrow
        \arg\max_{\substack{\mathcal Q\subseteq\mathcal C_{\mathrm{deg}}\\
        |\mathcal Q|\leq B_{\mathrm{review}}}}
        \sum_{k\in\mathcal Q}H_k$
    \ENDIF

    \STATE \textbf{// Replay-gated governance}
    \FOR{each distinct $k$ in periodic or budgeted event-driven review}
        \STATE Propose $z_k^{\mathrm{prop}}$ from qualified evidence:
        \textsc{Narrow}, \textsc{Repair}, or no proposal
        \STATE For feasible candidates, evaluate
        $\operatorname{Gate}(z_k^{\mathrm{prop}},k)$
        by Eq.~\ref{eq:replay-gate}
        \STATE $z_k^*\leftarrow$ Eq.~\ref{eq:action-selection};
        update the service state in $\mathcal K$
    \ENDFOR
\ENDFOR
\STATE \textbf{return} $(\mathcal M,\mathcal K)$
\end{algorithmic}
\end{algorithm}

\section{Method Implementation Details}
\label{app:method-details}

\subsection{Evidence Assignment and Skill Promotion}
\label{app:evidence-promotion}

When execution feedback provides object-level positive confidence, it determines $c^{+}_{k,u}$. With only task-level success, the $M_{t,s}$ distinct provenance families used in scope $s$ each receive $c^{+}_{k,u}=1/M_{t,s}$. A source memory and its derived skill form one family, with effective credit assigned only once. Failed trajectories remain BadCases until local feedback identifies a team-, event-, or agent-level object; unresolved uses produce $(y_{k,u},c_{k,u})=(\emptyset,0)$.

Procedural memory $m$ tracks recent effectiveness with
\[
\operatorname{EMA}_{m,j}
=
\rho\operatorname{EMA}_{m,j-1}+(1-\rho)x_{m,j},
\]
where $j$ indexes admitted observations, $x_{m,j}=1$ for positive evidence, and $x_{m,j}=0$ for qualified negative evidence. Unresolved outcomes do not update the sequence. Define
\begin{align}
A_m^{\mathrm{ready}}
&=1+n_m^{+}\operatorname{EMA}_m,
\qquad
B_m^{\mathrm{ready}}
=1+n_m^{-}(2-\operatorname{EMA}_m), \nonumber\\
\operatorname{Ready}(m)
&=\mu(A_m^{\mathrm{ready}},B_m^{\mathrm{ready}})
-\sigma(A_m^{\mathrm{ready}},B_m^{\mathrm{ready}}).
\label{eq:readiness}
\end{align}
Here $n_m^{+}$ and $n_m^{-}$ are the numbers of admitted positive and negative observations. A seed is eligible for promotion only if
\begin{equation}
\label{eq:promotion-gate}
\operatorname{Ready}(m_{\mathrm{seed}})
>
\frac{1}{|\mathcal P|}
\sum_{m\in\mathcal P}\operatorname{Ready}(m)
+\delta_{\mathrm{prom}},
\end{equation}
where $\mathcal P$ contains active memories from the same scope and, at agent level, the same identity; we set $\delta_{\mathrm{prom}}=0.05$. After promotion, the fast inherited mass in Equation~\ref{eq:evidence-transfer} is discounted with $\lambda=1-1/W$. Direct skill-use counts remain zero, and an equivalent existing skill is linked without migrating the same evidence again.

\subsection{Beta-Style Belief Drift and Uncertainty}
\label{app:reliability-details}

Equation~\ref{eq:mean-drift} provides an interpretable mean-level degradation signal. To account for Beta-style belief uncertainty in implementation, independently draw paired samples $\theta_{k,j}^{S}$ and $\theta_{k,j}^{F}$ from the slow and fast beliefs, and define $\Delta_{k,j}=\theta_{k,j}^{S}-\theta_{k,j}^{F}$ and $\bar{\Delta}_k=\mu_k^{S}-\mu_k^{F}$. AECG subtracts the centered sampling-noise baseline:
\begin{equation}
R_k^{\mathrm{drift}}
=
\left[
\frac{1}{J}\sum_{j=1}^{J}
\left(
[\Delta_{k,j}]_+
-
[\Delta_{k,j}-\bar{\Delta}_k]_+
\right)
\right]_+ .
\label{eq:drift}
\end{equation}

The following proposition characterizes the population quantity estimated by Equation~\ref{eq:drift}. It shows that the baseline subtraction is equivalently an expectation under the Beta-style beliefs of clipped positive degradation, rather than a calibrated probability or hypothesis-test statistic.

\paragraph{Proposition 1 (Expected clipped degradation).}
Let
\[
X_k:=\theta_k^{S}-\theta_k^{F},
\qquad
\delta_k:=\mathbb E[X_k]=\mu_k^{S}-\mu_k^{F},
\]
under the independent belief-sampling law used in Equation~\ref{eq:drift}, and define its population counterpart
\begin{equation}
\label{eq:population-drift}
\mathcal R_k^{\mathrm{drift}}
:=
\left[
\mathbb E\!\left([X_k]_+-[X_k-\delta_k]_+\right)
\right]_+.
\end{equation}
Then
\begin{equation}
\label{eq:clipped-drift}
\mathcal R_k^{\mathrm{drift}}
=
\begin{cases}
0, & \delta_k\leq 0,\\[2mm]
\mathbb E\!\left[\min\{\max\{X_k,0\},\delta_k\}\right],
& \delta_k>0,
\end{cases}
\end{equation}
and, for $\delta_k>0$,
\begin{equation}
\label{eq:integrated-exceedance}
\mathcal R_k^{\mathrm{drift}}
=
\int_{0}^{\delta_k}
\Pr(X_k>t)\,dt.
\end{equation}
Consequently,
\begin{equation}
\label{eq:drift-bounds}
0
\leq
\mathcal R_k^{\mathrm{drift}}
\leq
[\delta_k]_+,
\qquad
\delta_k\Pr(X_k\geq\delta_k)
\leq
\mathcal R_k^{\mathrm{drift}}
\leq
\delta_k\Pr(X_k>0)
\quad(\delta_k>0).
\end{equation}
Thus, the score is bounded by the belief-mean degradation and attenuates that degradation when belief samples do not consistently support $\theta_k^{S}>\theta_k^{F}$. Equivalently, $\mathcal R_k^{\mathrm{drift}}/\delta_k$ is the average belief-based exceedance score over thresholds $t\in[0,\delta_k]$.

\paragraph{Proof.}
Let $h(x)=[x]_+$. If $\delta_k\leq 0$, then $X_k-\delta_k\geq X_k$ pointwise. Since $h$ is nondecreasing,
\[
h(X_k)-h(X_k-\delta_k)\leq 0,
\]
and the outer positive part in Equation~\ref{eq:population-drift} gives $\mathcal R_k^{\mathrm{drift}}=0$.

Suppose instead that $\delta_k>0$. A pointwise case analysis gives
\[
h(x)-h(x-\delta_k)
=
\begin{cases}
0, & x\leq 0,\\
x, & 0<x<\delta_k,\\
\delta_k, & x\geq\delta_k,
\end{cases}
=
\min\{\max\{x,0\},\delta_k\}.
\]
Taking expectations proves Equation~\ref{eq:clipped-drift}; the integrand is nonnegative, so the outer positive part is redundant in this case. The layer-cake identity for a nonnegative random variable bounded by $\delta_k$ then yields
\[
\mathbb E\!\left[\min\{\max\{X_k,0\},\delta_k\}\right]
=
\int_0^{\delta_k}
\Pr(X_k>t)\,dt,
\]
which proves Equation~\ref{eq:integrated-exceedance}. The first pair of bounds in Equation~\ref{eq:drift-bounds} follows immediately from
$0\leq\min\{\max\{X_k,0\},\delta_k\}\leq\delta_k$.
For $\delta_k>0$, the pointwise inequalities
\[
\delta_k\mathbf 1\{X_k\geq\delta_k\}
\leq
\min\{\max\{X_k,0\},\delta_k\}
\leq
\delta_k\mathbf 1\{X_k>0\}
\]
give the remaining bounds after taking expectations.
\hfill$\square$

Because each summand in Equation~\ref{eq:drift} is bounded, the strong law of large numbers implies that its Monte Carlo estimate converges almost surely to $\mathcal R_k^{\mathrm{drift}}$ as $J\rightarrow\infty$. Moreover, when the sampled belief difference concentrates at a positive value $\delta_k$, dominated convergence gives $\mathcal R_k^{\mathrm{drift}}\rightarrow\delta_k$, recovering the mean-level degradation in Equation~\ref{eq:mean-drift}. These properties motivate the score as an uncertainty-adjusted degradation magnitude; they do not imply universal superiority over exceedance scores or belief-quantile detectors. We therefore interpret it as a ranking score rather than as a calibrated test statistic.

We use $J=1024$ with a deterministic seed derived from the belief snapshot. Fast-belief uncertainty is recorded as
\begin{equation}
\label{eq:uncertainty}
U_k=
\frac{\alpha_k^{F}\beta_k^{F}}
{(\alpha_k^{F}+\beta_k^{F})^2
(\alpha_k^{F}+\beta_k^{F}+1)}.
\end{equation}
It is retained for audit rather than mixed with hand-set risk weights.

\subsection{Triage and Action-Specific Validation}
\label{app:governance-details}

In Equation~\ref{eq:observed-exposure}, $\leadsto$ is a nonempty directed
path and converging paths are deduplicated by downstream node. Immediate
dependents receive unit weight, while distant decisions receive geometrically
smaller weight; at $\gamma=1$, exposure reduces to descendant counting. This
graph-based quantity upper-bounds potential propagation and does not claim
that every reachable node was causally changed. Distance discounting also
prevents graph depth alone from making high-level skills dominate. We set
$\gamma=0.5$ and use same-scope medians when exposure history is missing.

From the paired samples in Equation~\ref{eq:drift}, the confidence gate is
\begin{equation}
\label{eq:degradation-confidence}
C_k^{\mathrm{deg}}
:=
\frac{1}{J}\sum_{j=1}^{J}\mathbf 1\{\Delta_{k,j}>0\}.
\end{equation}
This Monte Carlo fraction is a confidence score under the Beta-style beliefs,
not a calibrated probability of degradation. With
$\eta_{\mathrm{deg}}=0.95$, let $\mathcal C_{\mathrm{deg}}$ contain skills
satisfying $C_k^{\mathrm{deg}}\geq\eta_{\mathrm{deg}}$ and
$R_k^{\mathrm{drift}}>0$. The periodic review set is
\begin{equation}
\label{eq:review-selection}
\mathcal Q^*
:=
\arg\max_{\substack{\mathcal Q\subseteq\mathcal C_{\mathrm{deg}}\\
|\mathcal Q|\leq B_{\mathrm{review}}}}
\sum_{k\in\mathcal Q}H_k.
\end{equation}
Thus, review selects the top $B_{\mathrm{review}}$ eligible severities.
$B_{\mathrm{review}}$ is fixed across methods, while qualified local failures
use a reserved event-driven budget.

For \textsc{Narrow}, a constrained generator produces a schema-valid
predicate over task, tool, event, agent, and input features that excludes the
implicated context without changing the procedure. For \textsc{Repair}, it
preserves scope and intent while revising the procedure or its avoidance
rules.

Failure fixtures contain qualified same-scope negative cases; disjoint
regression fixtures sample prior successful uses. \textsc{Narrow} withholds
predicate-matched uses and runs regression replay outside the excluded region,
whereas \textsc{Repair} runs the revision on both sets. Missing either set
makes the candidate ineligible. We set $\epsilon_{\mathrm{reg}}=0.05$.

For fallback, define
\[
C_k^{\mathrm{unsafe}}
:=
F_{\operatorname{Beta}(\alpha_k^F,\beta_k^F)}(1/2).
\]
This fast-belief confidence gate is not a calibrated probability that the
skill is unsafe. The final action is
\begin{equation}
\label{eq:action-selection}
z_k^*
:=
\begin{cases}
z_k^{\mathrm{prop}},&
z_k^{\mathrm{prop}}\in\{\textsc{Narrow},\textsc{Repair}\},
\ \operatorname{Gate}(z_k^{\mathrm{prop}},k)=1,\\
\textsc{Isolate},&
C_k^{\mathrm{unsafe}}\geq\eta_{\mathrm{unsafe}},\\
\textsc{Keep},&\text{otherwise},
\end{cases}
\end{equation}
with $\eta_{\mathrm{unsafe}}=0.95$. Only candidates that pass paired
replay become active versions. If no candidate is generated, either
fixture set is missing, or replay fails, the parent version is isolated
when $C_k^{\mathrm{unsafe}}\geq\eta_{\mathrm{unsafe}}$; otherwise, it
remains active under monitoring. Any unvalidated candidate remains
inactive.

\subsection{Hyperparameter Selection}
\label{app:hyperparameters}

To ensure fair evaluation and prevent test-set overfitting, we do not tune hyperparameters on the evaluation benchmarks. Reliability and promotion parameters ($\kappa$, $W$, $\lambda$, and $\delta_{\mathrm{prom}}$) were selected using a small, independent hold-out development set constructed from a subset of AgentBoard interactive tasks not used in our main evaluation. Governance uses fixed design values $\gamma=0.5$, $\eta_{\mathrm{deg}}=\eta_{\mathrm{unsafe}}=0.95$, and $\epsilon_{\mathrm{reg}}=0.05$; $B_{\mathrm{review}}$ is a deployment budget held equal across compared methods rather than a benchmark-tuned accuracy parameter.

Specifically, we perform a grid search on the development set:
\begin{itemize}
    \item \textbf{Evidence Inheritance ($\kappa$):} We tested $\kappa \in \{0.3, 0.5, 0.7, 1.0\}$. We set $\kappa = 0.5$ as it optimally balances the preservation of historical confidence with the epistemic uncertainty introduced by LLM-based abstraction.
    \item \textbf{Discount Window ($W$) and Factor ($\lambda$):} We searched $W \in \{5, 10, 20\}$. We selected $W = 10$, which mathematically couples with the fast discount factor $\lambda = 1 - 1/W = 0.9$. This configuration ensures that recent performance degradation ($D_k$) is statistically detectable within 3-5 consecutive failures, allowing timely governance interventions.
    \item \textbf{Promotion Margin ($\delta_{\mathrm{prom}}$):} This operation threshold prevents premature skill abstraction. We empirically set $\delta_{\mathrm{prom}} = 0.05$, which prevents spurious trajectory memorization while ensuring that consistently useful procedures are promoted efficiently.
\end{itemize}
Importantly, these parameters are kept fixed across all three multi-agent frameworks (AutoGen, MacNet, AgentNet) and all four evaluation benchmarks (PDDL, FEVER, ScienceWorld, MBPP-Plus), demonstrating the generalizability and robustness of the AECG framework without requiring per-task tuning.
\subsection{LLM-based Failure Attribution}
\label{app:evidence-attribution}

AECG uses LLM-based attribution only to determine whether a failed execution
provides sufficiently qualified negative evidence for a previously applied
skill. The attribution component is not treated as a source of ground-truth
causal labels.

\paragraph{Applied-skill records.}
During execution, the memory manager records the skills retrieved, selected,
and injected into each agent context, together with their coordination
scope, execution node, and downstream nodes. When the agent explicitly
reports the skills it used, this report is used as the applied-skill record.
For agents without structured skill-use reporting, AECG falls back to a
compatibility mode in which injected skills are treated as applied and the
record is marked as
\texttt{compat\_assume\_injected}. Thus, each applied-skill record is either
explicitly reported or conservatively inferred from the injected context.

\paragraph{Failure-case construction.}
When a task fails, AECG constructs a \emph{BadCase} containing the task
specification and input, the final agent output, authoritative environment
feedback, the execution trace, and the applied-skill records. Each candidate
skill is therefore grounded in an observed execution record rather than
introduced by the attribution model.

\paragraph{Dual-judge attribution.}
Two separately sampled LLM judges independently inspect the BadCase. Each
judge outputs a responsible scope, responsible execution node, implicated
skill identifier, failure type, confidence score, and an optional abstention
decision. Judges are restricted to skills present in the applied-skill
record and cannot introduce new skill identifiers.

Let the two judge outputs for failed task $t$ be
\[
J_i(t)
=
(\widehat{s}_i,\widehat{v}_i,\widehat{k}_i,
\widehat{f}_i,\widehat{c}_i,a_i),
\qquad i\in\{1,2\},
\]
where $\widehat{s}_i$ denotes the predicted scope,
$\widehat{v}_i$ the responsible execution node,
$\widehat{k}_i$ the implicated skill,
$\widehat{f}_i$ the predicted failure type,
$\widehat{c}_i\in[0,1]$ the judge confidence, and
$a_i\in\{0,1\}$ indicates whether the judge abstains.

\paragraph{Attribution gate.}
A failed observation is admitted as negative evidence for skill $k$ only if
all of the following conditions hold:
(i) neither judge abstains;
(ii) both judges identify the same skill;
(iii) both judges identify the same scope;
(iv) the identified skill is present in the applied-skill record; and
(v) the mean judge confidence is at least $0.70$.

Formally, defining
\[
\overline{c}_{k,t}
=
\frac{\widehat{c}_1+\widehat{c}_2}{2},
\]
the attribution gate is
\begin{equation}
\label{eq:attribution-gate}
G_{k,t}
=
\mathbb{I}\!\left[
a_1=a_2=0
\land
\widehat{k}_1=\widehat{k}_2=k
\land
\widehat{s}_1=\widehat{s}_2
\land
k\in\mathcal{K}^{\mathrm{applied}}_t
\land
\overline{c}_{k,t}\geq 0.70
\right].
\end{equation}

If $G_{k,t}=1$, the failure contributes
$y_{k,u}=0$ with confidence
$c^{\mathrm{att}}_{k,u}=\overline{c}_{k,t}$.
Otherwise, the observation is marked unresolved,
$y_{k,u}=\emptyset$, and no negative evidence is admitted.

This gate deliberately favors abstention over uncertain attribution. In
particular, disagreement between judges, explicit abstention, or insufficient
confidence cannot directly decrease the reliability belief of a skill.

\section{Experimental Details}
\label{app:experimental-details}

\subsection{Benchmarks, Baselines, Frameworks, and Models}
\label{app:benchmarks-baselines}

\paragraph{Benchmarks.}
PDDL, adopted from AgentBoard \citep{ma2024agentboard}, evaluates interactive
planning by task success rate. FEVER \citep{thorne2018fever} evaluates
evidence-based fact verification by classification accuracy. ScienceWorld
\citep{wang2022scienceworld} tests multi-step scientific reasoning and action
in a text environment. MBPP-Plus from EvalPlus \citep{liu2023your}
evaluates Python synthesis using augmented hidden tests. Together they span
planning, verification, embodied interaction, and code generation.

\paragraph{Baselines.}
No-Memory removes cross-task persistent memory while retaining native
within-task context. MacNet-M retains MacNet's lightweight artifact memory
\citep{qian2025scaling}. MemoryBank uses long-term storage with
Ebbinghaus-inspired forgetting and reinforcement
\citep{zhong2024memorybank}. ConMem coordinates useful strategies and failure
cues as signed cards in a relational graph \citep{tan2026conmem}. G-Memory
organizes insights and condensed collaboration trajectories hierarchically
\citep{zhang2026g}.

\paragraph{Frameworks and controls.}
AutoGen \citep{wu2023autogen} is conversation-centric, MacNet
\citep{qian2025scaling} organizes collaboration as a directed acyclic graph,
and AgentNet \citep{yang2026agentnet} supports decentralized routing. For
every framework--benchmark pair, methods share task instances, prompts,
tools, execution limits, evaluators, and memory budgets. Agent execution uses
\texttt{gpt-4.1-mini}; trajectory consolidation, skill abstraction,
memory-side assessment, and intervention proposals use
\texttt{gpt-4o-mini}. Each configuration is evaluated in three independently
seeded runs, and tables report arithmetic means.

\subsection{Additional Main-Result Analysis}
\label{app:main-results-extra}

\begin{figure}[t]
\centering
\includegraphics[width=\textwidth,trim=0 0 0 0,clip]{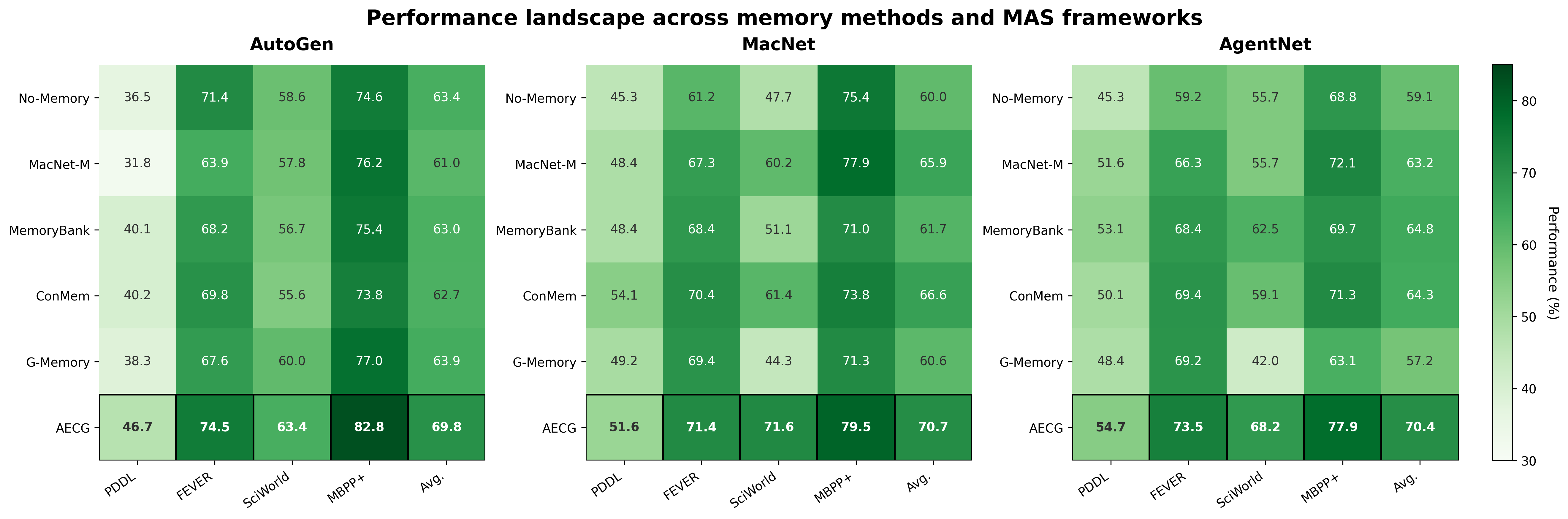}
\caption{Performance landscape across frameworks, benchmarks, and memory
methods. The outlined AECG row corresponds to Table~\ref{tab:main-results}.}
\label{fig:main-results-heatmap}
\end{figure}

Figure~\ref{fig:main-results-heatmap} highlights two benchmark-wise patterns.
First, the largest improvements occur on ScienceWorld and MBPP-Plus. Under
MacNet, AECG raises ScienceWorld accuracy from 47.73 for No-Memory to 71.59
and MBPP-Plus from 75.41 to 79.51; under AgentNet, the corresponding gains are
55.68 to 68.18 and 68.85 to 77.87. These tasks repeatedly reuse intermediate
procedures and are especially sensitive to stale or over-generalized skills.
Second, PDDL is the most competitive benchmark: ConMem slightly outperforms
AECG on MacNet, although AECG remains best on AutoGen and AgentNet. Thus, the
overall advantage is not attributable to one framework or task family.

\subsection{Mechanism-Test Protocols}
\label{app:mechanism-protocols}

\paragraph{Shared protocol.}
The controlled-corruption and budgeted-triage experiments use AgentNet--ScienceWorld because its execution graph exposes
agent-, event-, and team-level decisions and their downstream dependencies.
For each seed, the first 60\% of the task stream warms memory and the remaining
40\% evaluates it. Methods start from the same post-warm-up snapshot and share
task order, models, prompts, tools, retrieval limits, and execution budgets.

\paragraph{Controlled corruption.}
After warm-up, we select four active skills: two with relatively high predicted
exposure and two with relatively low exposure. Each has at least three positive
observations and a fast-belief mean of at least 0.8. An LLM changes one
critical step or applicability condition while preserving scope and principal
retrieval cues. On three matched validation tasks, the original skill must
succeed at least twice, the corruption must fail at least twice, and the
corrupted skill must still be retrieved at least twice. We inject the same
validated corruptions into a shared snapshot. Detection delay counts skill
uses from the first failed use to first review; polluted-skill failed uses
count skill-use events in failed tasks and can exceed the number of tasks.

\paragraph{Scope mechanism.}
Hard-scope AECG requires source and decision scopes to match, including agent
identity for agent-level objects. Flat-scope removes this eligibility
constraint while retaining the same top-$K$ and token budget. Cross-scope use
rate is the fraction of selected uses whose source and decision scopes differ.
Failed-task potential exposure averages their distance-discounted observed
exposure in failed tasks and measures potential reach, not causal propagation.

\paragraph{Budgeted triage.}
We reuse controlled corruption and allow at most one review per round; the
number of confidence-qualified candidates exceeds this budget in the first
two rounds of every run. We compare $E_kR_k^{\mathrm{drift}}$,
$R_k^{\mathrm{drift}}$, and random qualified selection, isolating every
selected skill to hold intervention fixed. Coverage@2 is the fraction of total
polluted-skill predicted exposure covered by the first two reviews and is a
ranking diagnostic. Post-corruption success is the primary outcome, while
realized failed-task exposure is computed from post-intervention traces.

\section{Prompt Templates}
\label{app:prompt}

The following excerpts are abridged versions of the prompts used by AECG. We
retain the constraints that determine the behavior of skill formation, failure
attribution, evidence-routed intervention, and replay-gated revision.
Formatting instructions, few-shot examples, and implementation-specific
boilerplate are omitted.

\subsection{Skill Abstraction}
\label{app:prompt-skill-abstraction}

\begin{tcolorbox}[
  enhanced,
  breakable,
  colback=white,
  colframe=gray!90!black,
  colbacktitle=gray!90!black,
  coltitle=white,
  fonttitle=\bfseries\large,
  fontupper=\normalsize,
  title={SKILL ABSTRACTION PROMPT},
  arc=1mm,
  outer arc=1mm,
  boxrule=2.0pt,
  left=3mm, right=3mm,
  top=1mm, bottom=1mm,
  toptitle=-1mm,
  bottomtitle=-1mm
]
\textbf{Input.} A seed procedural memory, its execution trajectory and
reliability statistics, up to three auxiliary memories from the same
procedural level, and the target skill granularity (\texttt{team},
\texttt{event}, or \texttt{agent}).

\medskip
\textbf{Prompt excerpt.}

\medskip
You are the Skill Abstraction Generator in a continual-learning system.
Produce a reusable skill from the supplied seed memory.

The seed is the primary evidence. Its procedure must be an abstraction of
actions actually present in the seed; do not invent new knowledge, overturn
the seed's approach, or import actions that occur only in auxiliary memories.
Auxiliary memories may supplement applicability boundaries, risk limits, and
stable phrasing, but they are not primary evidence.

Preserve the coordination scope and match the granularity of the skill:
\begin{itemize}
    \item At the team level, describe how events are decomposed and handed off.
    \item At the event level, describe how agents divide work and combine outputs.
    \item At the agent level, describe tool calls, judgments, validation, and
    fallback behavior for the assigned role.
\end{itemize}

Remove concrete entities, runtime identifiers, timestamps, and task-specific
answers. State reusable applicability conditions and actionable avoidance
rules. Every avoidance rule must identify the failure scenario, the resulting
risk, and the behavior that should therefore be avoided.

Return exactly the following six fields:
\texttt{skill\_topic}, \texttt{applicable\_context},
\texttt{trigger\_condition}, \texttt{procedure}, \texttt{avoidance}, and
\texttt{evidence\_summary}. The evidence summary must state that the seed is
primary evidence and auxiliary memories are supplementary.
\end{tcolorbox}

\subsection{Conservative Failure Attribution}
\label{app:prompt-failure-attribution}

\begin{tcolorbox}[
  enhanced,
  breakable,
  colback=white,
  colframe=gray!90!black,
  colbacktitle=gray!90!black,
  coltitle=white,
  fonttitle=\bfseries\large,
  fontupper=\normalsize,
  title={CONSERVATIVE FAILURE ATTRIBUTION PROMPT},
  arc=1mm,
  outer arc=1mm,
  boxrule=2.0pt,
  left=3mm, right=3mm,
  top=1mm, bottom=1mm,
  toptitle=-1mm,
  bottomtitle=-1mm
]
\textbf{Input.} The task goal and outcome, the complete execution trace in
temporal order, applied-skill records grouped by exact scope, and the full
text of each skill that entered the corresponding scope.

\medskip
\textbf{Prompt excerpt.}

\medskip
You are one of two independent failure-attribution judges. Determine whether
an actually applied skill contributed to the observed failure. Do not infer
causality from application, topical similarity, or scope provenance alone.

First identify the observed failure and locate the earliest
evidence-supported execution node that deviated from a correct path. For every
applied skill, compare its trigger condition, procedure, and avoidance rules
with the concrete execution trace. Link a skill to the failure only when a
specific clause corresponds to an observed behavior and that behavior
participates in the failure chain.

Test competing explanations, including agent noncompliance, planning or
coordination errors, bad retrieval, tool or environment failures,
insufficient validation, and failures that began before skill application. An
applied skill is not proof that the agent followed it.

Recommend at most one exact \texttt{scope} and \texttt{skill\_id}. Abstain
when the trace is incomplete, the first deviation cannot be located, multiple
candidates are indistinguishable, or a stronger non-skill explanation
remains. Prefer conservative abstention over unsupported negative evidence.

Return a structured assessment containing the observed failure, the first
deviation, candidate assessments, clause links, alternative explanations,
responsibility classification, defect type, failure locality, and an
abstention decision. Use only clauses present in the supplied skill text.

The two judges are called independently. Negative evidence is admitted only
when both judges identify the same applied skill and scope, neither abstains,
and their agreement satisfies the confidence gate.
\end{tcolorbox}

\subsection{Context Narrowing}
\label{app:prompt-context-narrowing}

\begin{tcolorbox}[
  enhanced,
  breakable,
  colback=white,
  colframe=gray!90!black,
  colbacktitle=gray!90!black,
  coltitle=white,
  fonttitle=\bfseries\large,
  fontupper=\normalsize,
  title={CONTEXT NARROWING PROMPT},
  arc=1mm,
  outer arc=1mm,
  boxrule=2.0pt,
  left=3mm, right=3mm,
  top=1mm, bottom=1mm,
  toptitle=-1mm,
  bottomtitle=-1mm
]
\textbf{Stage A: Failure-Context Predicate}

\textbf{Input.} The current skill, attributed bad cases, and the context
snapshots recorded at the skill's actual application scope.

\medskip
\textbf{Prompt excerpt.}

\medskip
Propose the smallest executable predicate describing the failure context in
which the current skill should be excluded. Use only features present in the
supplied pre-execution context snapshots and supported by the attributed bad
cases. Do not infer features from identifiers, wording, final outcomes, or
unsupported domain knowledge.

The predicate uses the implemented matcher semantics: non-empty dimensions
are combined with AND; values within a dimension use the specified OR
semantics; agent composition requires exact set equality; and feature
constraints are conjunctive. Do not use an OR combination that expands beyond
the evidence.

Minimize the predicate. Retain a condition only if it distinguishes the
failure region, is observable before execution, and can be extracted
reliably. Abstain when the context snapshot is missing, the failure regions
cannot be represented by one safe predicate, or every non-empty predicate
would over-exclude valid uses.

Return either a versioned predicate with its supporting bad cases or an
abstention decision. Do not claim that replay has passed.

\medskip
\textbf{Stage B: Skill-Boundary Revision}

\textbf{Input.} The current skill, attributed bad cases, and the validated
failure-context predicate.

\medskip
\textbf{Prompt excerpt.}

\medskip
Revise only \texttt{applicable\_context} and \texttt{trigger\_condition}.
Exclude the predicate-defined failure region while preserving the original
skill's supported behavior elsewhere.

Treat the validated predicate as the primary boundary evidence. Use only
conditions observable before execution. Do not modify, restate, or supplement
the skill's \texttt{procedure} or \texttt{avoidance}, and do not disable the
entire skill unless the evidence covers its full original applicability.

Return the revised applicability and trigger conditions together with a
concise rationale identifying what was preserved, what was excluded, and
which evidence supports the boundary. The candidate remains inactive until
paired replay validates it.
\end{tcolorbox}

\subsection{Procedural Repair}
\label{app:prompt-procedural-repair}

\begin{tcolorbox}[
  enhanced,
  breakable,
  colback=white,
  colframe=gray!90!black,
  colbacktitle=gray!90!black,
  coltitle=white,
  fonttitle=\bfseries\large,
  fontupper=\normalsize,
  title={PROCEDURAL REPAIR PROMPT},
  arc=1mm,
  outer arc=1mm,
  boxrule=2.0pt,
  left=3mm, right=3mm,
  top=1mm, bottom=1mm,
  toptitle=-1mm,
  bottomtitle=-1mm
]
\textbf{Input.} The current skill, attributed bad cases, the first deviation,
clause-level attribution evidence, and relevant counterevidence.

\medskip
\textbf{Prompt excerpt.}

\medskip
Revise only \texttt{procedure} and \texttt{avoidance} to repair the smallest
concrete skill defect supported by the attribution evidence.

Distinguish an incorrect instruction, an ambiguous instruction, a missing
guardrail, and a wrong avoidance rule. Preserve useful steps, ordering, scope,
and intent that are not contradicted by the evidence. Make the revised
procedure executable: state branch conditions explicitly, verify critical
facts, validate critical outputs, and add stop, retry, fallback, or escalation
behavior only when supported by the failure evidence.

Do not change the applicability or trigger conditions. Do not hard-code the
answer to one bad case, invent tools or evidence, or introduce an entirely
new method. Every avoidance rule must name a concrete dangerous behavior and,
where necessary, state the replacement behavior.

Return the revised \texttt{procedure}, revised \texttt{avoidance}, and an
evidence-based rationale. Do not claim that replay has passed; the revision is
only a candidate until failure-fixture and regression replay succeed.
\end{tcolorbox}

\subsection{Replay Contract}
\label{app:prompt-replay-contract}

\begin{tcolorbox}[
  enhanced,
  breakable,
  colback=white,
  colframe=gray!90!black,
  colbacktitle=gray!90!black,
  coltitle=white,
  fonttitle=\bfseries\large,
  fontupper=\normalsize,
  title={REPLAY CONTRACT PROMPT},
  arc=1mm,
  outer arc=1mm,
  boxrule=2.0pt,
  left=3mm, right=3mm,
  top=1mm, bottom=1mm,
  toptitle=-1mm,
  bottomtitle=-1mm
]
\textbf{Input.} A candidate skill version, its implicated failure fixtures,
disjoint historical-success fixtures, and the deterministic replay and
governance policy.

\medskip
\textbf{Prompt excerpt.}

\medskip
Treat every generated excerpt above as producing a candidate skill version,
not as directly modifying an active skill. A candidate is activated only when
it mitigates the implicated behavior on failure fixtures and preserves
supported behavior on disjoint historical-success fixtures within the
regression tolerance.

Otherwise, keep the parent version active, isolate the candidate, or continue
monitoring according to the deterministic governance policy. Candidate
generation and replay must not be represented as evidence that the revision
has passed until both the failure and regression conditions are satisfied.
\end{tcolorbox}

\end{document}